\documentclass{ecai} 

\usepackage{latexsym}
\usepackage{amssymb}
\usepackage{amsmath}
\usepackage{amsthm}
\usepackage{booktabs}
\usepackage{enumitem}
\usepackage{graphicx}\usepackage{xcolor}
\usepackage{algorithmic}
\usepackage{algorithm}
\usepackage{caption}
\newtheorem{definition}{Definition}

\newcommand{\BibTeX}{B\kern-.05em{\sc i\kern-.025em b}\kern-.08em\TeX}

\begin{document}


\begin{frontmatter}


\paperid{123} 


\title{Verification-Guided Falsification for Safe RL via Explainable Abstraction and Risk-Aware Exploration}





\author[A]{\fnms{Tuan}~\snm{Le}\thanks{Corresponding Author. Email: tuanle@iastate.edu}}
\author[B]{\fnms{Risal}~\snm{Shefin}}
\author[B]{\fnms{Debashis}~\snm{Gupta}}
\author[C]{\fnms{Thai}~\snm{Le}}
\author[B]{\fnms{Sarra}~\snm{Alqahtani}}

\address[A]{Iowa State University}
\address[B]{Wake Forest University}
\address[C]{Indiana University}



\begin{abstract}
Ensuring the safety of reinforcement learning (RL) policies in high-stakes environments requires not only formal verification but also interpretability and targeted falsification—the deliberate search for counterexamples that expose potential failures prior to deployment. While model checking provides formal guarantees, its effectiveness is limited by abstraction quality and the completeness of the underlying trajectory dataset. We propose a hybrid framework that integrates: (1) explainability, (2) model checking, and (3) risk-guided falsification to achieve both rigor and coverage. Our approach begins by constructing a human-interpretable abstraction of the RL policy using Comprehensible Abstract Policy Summarization (CAPS). This abstract graph, derived from offline trajectories, is both verifier-friendly, semantically meaningful, and can be used as input to Storm probabilistic model checker to verify satisfaction of temporal safety specifications. If the model checker identifies a violation, it will return an interpretable counterexample trace by which the policy fails the safety requirement. However, if no violation is detected, we cannot conclude satisfaction due to potential limitation in the abstraction and coverage of the offline dataset. In such cases, we estimate associated risk during model checking to guide a falsification strategy that prioritizes searching in high-risk states and regions underrepresented in the trajectory dataset. We further provide PAC-style guarantees on the likelihood of uncovering undetected violations. Finally, we incorporate a lightweight safety shield that switches to a fallback policy at runtime when such a risk exceeds a threshold, facilitating failure mitigation without retraining.  Empirical results in safety-critical RL domains—including two tasks from the Mujoco benchmark and a medical task (insulin dosing)—demonstrate that our framework not only detects significantly more safety violations than uncertainty-based and fuzzy-based search methods, but also uncovers a more diverse and novel set of counterexamples. These counterexamples span a wider range of trajectories and failure modes, providing richer insights into the policy's limitations that can help debug, understand, and repair unsafe policies.
\end{abstract}

\end{frontmatter}


\section{Introduction}

Reinforcement Learning (RL) has demonstrated remarkable success, achieving superhuman performance in games and simulated environments. Despite these achievements, the deployment of RL in real-world, safety-critical domains—such as autonomous driving, robotics, and healthcare—remains limited. This gap is primarily due to two closely interrelated challenges: ensuring the \emph{safety} of RL policies and providing \emph{interpretable explanations} for their decisions. While substantial research has been devoted to these aspects individually, they are rarely addressed jointly. 

Research in \emph{safe RL} has traditionally focused on training agents that adhere to explicit safety constraints~\cite{kim2020safe, geibel2006reinforcement, recRL, bastani2021safe, alshiekh2018safe, mihatsch2002risk}. Meanwhile, \emph{explainable RL} (XRL) research emphasizes the transparency and human interpretability of agents' behavior~\cite{ruggeri2025explainable}. However, interpretability is more than a usability feature. It is also crucial for robust safety assurance. Without clear explanations, practitioners cannot reliably determine whether an agent complies with safety constraints, diagnose the causes of violations, or take timely interventions. This work argues that \emph{safety assurance and interpretability must be integrated}, and we propose that interpretable policy abstractions can serve as a powerful foundation for formal safety verification.

However, formally verifying the safety of RL policies poses several key challenges. First, RL typically involves extremely large or continuous state spaces, creating state explosion problems that render the direct construction of complete Markov Decision Process (MDP) models computationally infeasible~\cite{corsi2021formal}. Second, unlike traditional verification contexts where system dynamics are known explicitly, RL policies are inherently learned from data and often represented as opaque, black-box functions, necessitating abstraction methods that can extract interpretable, verifiable models from empirical data~\cite{cheng2025survey}. Third, RL policies are typically trained using finite datasets or a limited number of simulation rollouts, leaving critical parts of the state space underrepresented or unexplored, resulting in blind spots during verification~\cite{ruggeri2025explainable}. Fourth, when safety violations occur, practitioners require intuitive, human-readable explanations that facilitate effective debugging and remediation. Yet interpretability in the context of RL verification remains underdeveloped. Finally, many existing safety approaches in RL are reactive, often necessitating extensive retraining or significant redesign to address localized failures~\cite{recRL, bastani2021safe}, analogous to rewriting an entire software module due to a single failing test.

Existing work on RL safety falsification—defined as the deliberate search for counterexamples that expose policy violations of safety specifications—typically falls into two distinct categories, each with significant limitations. Search-based testing approaches~\cite{karakovskiy2012mario, manes2021art} leverage exploration strategies such as depth-first search or fuzzing to reveal policy weaknesses. However, they lack formal safety guarantees and are computationally expensive due to extensive simulation requirements. Conversely, formal verification methods, especially those employing probabilistic model checking techniques such as Storm~\cite{hahn2019storm, wang2020towards}, provide rigorous guarantees via temporal logic specifications (e.g., Probabilistic Computation Tree Logic, PCTL), yet often suffer from abstraction-induced blind spots. To the best of our knowledge, no previous work has integrated falsification, interpretability, and formal verification into a single cohesive approach capable of addressing these challenges simultaneously.

In this work, we present a hybrid approach that combines formal verification with targeted falsification, using interpretable policy abstraction as a bridge between explainability and verification. We begin by summarizing the RL policy into a directed graph using Comprehensible Abstract Policy Summarization (CAPS) \cite{McCalmon}, an XRL technique that generates human-interpretable and verifier-friendly representations. CAPS graph, constructed from offline trajectories, serves as input to the Storm model checker, which verifies probabilistic temporal logic specifications (e.g., PCTL). If a safety violation is found, we conclude with certainty that the policy is unsafe. However, if no violation is detected, we cannot assume the policy is safe, since the abstraction may omit critical transitions and the offline dataset may lack full coverage. To address this, we use risk estimates obtained during model checking to guide a falsification strategy that actively explores high-risk and underrepresented regions of the state space. Specifically, this strategy combines risk critics and ensemble-based epistemic uncertainty to identify violations that formal verification might miss.


\textbf{Contributions.} Our novel framework for safety assurance of RL policies integrates (1) explainable abstraction, (2) formal model checking, and (3) risk-guided falsification. Our contributions are:

\vspace{-5pt}
\begin{enumerate}
    \item \textbf{Explainable Abstraction and Model Checking:} We use CAPS to abstract the RL policy into an interpretable graph and verify it using the Storm model checker. This enables formal safety guarantees with transparent counterexamples.
    
    \item \textbf{Risk- and Uncertainty-Guided Falsification:} When model checking cannot confirm safety, we invoke a targeted falsification strategy guided by risk estimates from the model checking and ensemble-based uncertainty, focusing on high-risk and poorly represented states.

    \item \textbf{PAC-Style Safety Guarantees:} We provide $(\epsilon_r, \delta)$-PAC guarantees on the completeness of safety violation detection under abstraction and data limitations.

    \item \textbf{Runtime Safety Shielding:} We introduce a lightweight shielding mechanism that switches to a fallback policy in high-risk states, enabling modular policy repair without retraining.
\end{enumerate}
\vspace{-5pt}

We evaluate our framework extensively across benchmark domains, including CMDB Mujoco Navigation2, Maze, and a simulated medical task for regulating insulin dosing in type 2 diabetes patients. Empirical results show that our approach significantly outperforms baseline verification methods, detecting substantially more safety violations while enhancing interpretability and enabling modular runtime interventions. By unifying explainable policy abstraction, formal verification, and guided falsification, this work advances the state of the art in RL safety assurance, paving the way for practical and reliable deployment in high-stakes environments.

\section{Background}
\noindent \textbf{Related Work. }
Safety assurance in RL has been approached through both formal verification and falsification techniques, each addressing complementary aspects of the problem. Formal verification in RL often leverages probabilistic model checking tools such as PRISM~\cite{kwiatkowska2011prism} and Storm~\cite{dehnert2017storm} that provide rigorous guarantees by verifying temporal logic properties (e.g., PCTL) over Markov decision processes. However, applying these tools to RL is challenging due to the large or continuous state spaces and the black-box nature of learned policies. To enable tractable verification, prior work has explored abstraction techniques, including state aggregation and symbolic representations~\cite{li2006towards}. However, they often sacrifice or do not consider model interpretability that is critical for understanding and mitigating safety violations.

Falsification methods offer a complementary strategy by actively searching for counterexamples that reveal unsafe behaviors. Such methods include random search, optimization-based falsification, and RL-guided falsifiers~\cite{deshmukh2015stochastic, karunakaran2020counterexample}. While these techniques are scalable and can expose vulnerabilities in complex systems, they lack formal guarantees and often require significant computational effort. To bridge the gap between exhaustive verification and scalable falsification, some recent approaches combine model checking with runtime shielding~\cite{alshiekh2018safe} by using formal tools to detect violations and then switching policies dynamically to avoid them. Nonetheless, existing methods tend to treat abstraction, interpretability, and falsification as separate concerns. In contrast, our work builds on these foundations by tightly integrating explainable policy abstraction, model checking, and risk-aware falsification into a single framework. Particularly, we use CAPS to generate a semantically meaningful abstraction from offline trajectories, enabling formal verification through Storm. When model checking fails to detect violations such as due to limited coverage or abstraction, we use the risk estimates from the abstract model to guide a targeted falsification process. This combination provides formal safety analysis with practical falsification capabilities while supporting interpretability throughout.

\vspace{5pt}
\noindent \textbf{Probabilistic Computation-tree Logic. }
Probabilistic Computation Tree Logic (PCTL) is a temporal logic for specifying probabilistic safety and reachability properties in stochastic systems~\cite{baier2008principles}. It extends CTL by enabling probability constraints over system behaviors, and is widely used in popular model checking tools like PRISM \cite{baier2008principles} and Storm \cite{hahn2019storm}. PCTL formulas follow the grammar:
\begin{align*}
\Phi ::= & \ a \mid \neg \Phi \mid \Phi \land \Phi \mid \mathbb{P}_{J}(\phi) \
\phi ::= & \ X \Phi \mid \Phi \ U\ \Phi \mid \Phi\ U^{\leq n}\ \Phi
\end{align*}

Here, $a$ is an atomic proposition; $\neg$ and $\land$ are logical negation and conjunction. $\mathbb{P}_{J}(\phi)$ means the path formula $\phi$ holds with probability in interval $J \subseteq [0,1]$. The main temporal operators are; $X \Phi$: $\Phi$ holds in the next state (Next),  $\Phi_1\ U\ \Phi_2$: $\Phi_1$ holds until $\Phi_2$ becomes true (Until), and $\Phi_1\ U^{\leq n}\ \Phi_2$: same as above but within $n$ steps (Bounded Until). For example, $\mathbb{P}_{\leq 0.01}(\ U^{\leq 100}\ \textit{unsafe})$ states that reaching an unsafe state within 100 steps has probability at most 0.01.

\begin{figure*}[h]
    \centering
    \includegraphics[width=0.95\textwidth]{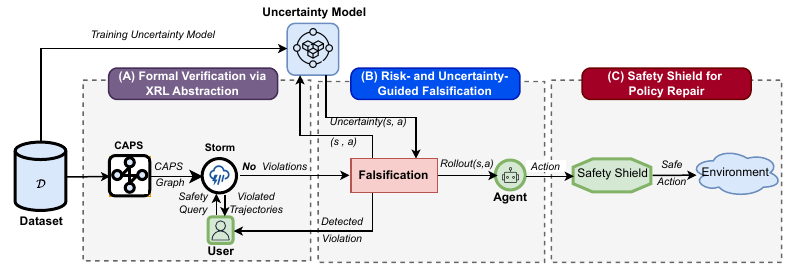}
    \caption{Our hybrid approach combines (A) formal verification, (B) risk-aware falsification, and (C) explainable shielding for RL policy safety.}
    \label{fig:overview}
    \vspace{5pt}
\end{figure*}

\section{Problem Setting}
\label{sec:problem_setting}

We consider the standard Constrained Markov Decision Process (CMDP), $\mathcal{M} = (\mathcal{S}, \mathcal{A}, \mu, \mathcal{P}, \mathcal{R}, \gamma, \mathcal{C}, T)$, defined by a state space $\mathcal{S}$, action space $\mathcal{A}$, initial distribution $\mu$, transition dynamics $\mathcal{P}$, reward function $\mathcal{R}$, discount factor $\gamma$, safety cost functions $\mathcal{C} = \{c_i: \mathcal{S} \times \mathcal{A} \to \mathbb{R}_{\geq 0}\}_{i=1}^k$, and finite horizon $T$. CMDPs naturally capture the dual objectives of maximizing reward while satisfying safety constraints, making them well-suited for formal verification.

We assume access to a fixed offline dataset  $\mathcal{D} = \{(s_i,a_i,s'_i,r_i,\mathbf{c}_i,d_i)\}_{i=1}^N$ containing historical interactions, with $\mathbf{c}_i = (c_i^1,\dots,c_i^k)$ denoting constraint violations and $d_i$ indicating episode termination. A limited budget of additional environment interactions may be allowed. To enable tractable verification, we abstract the policy $\pi_{\text{task}}$ into a directed graph $\mathcal{G}_\pi = (V, E)$, where nodes $V \subset \mathcal{S}$ represent states and edges $E \subseteq V \times V$ represent transitions under $\pi_{\text{task}}$, annotated with empirical transition probabilities and safety costs. Verification is then cast as a probabilistic model checking task on $\mathcal{G}_\pi$ against a temporal safety property $\varphi$ (e.g., $\mathbb{P}_{\geq 0.95}[\neg \text{unsafe} \ \mathcal{U}^{\leq T} \ \text{goal}]$). However, abstraction-based verification introduces two core challenges: (1) \textbf{Abstraction Error}—limited samples in $\mathcal{D}$ may omit critical transitions, yielding incomplete or misleading abstractions; and (2) \textbf{Epistemic Uncertainty}—sparse data leads to uncertainty in transition and cost estimates, particularly in underexplored regions. 

These issues differentiate our setting from traditional uncertainty estimation contexts, such as active learning in supervised learning~\cite{settlesactivelearninginpractice, cohn1996activelearning, bachman2017learningalgforactivelearning}, where uncertainty typically relates to predictive error over a static input distribution. In contrast, uncertainty in CMDPs reflects incomplete knowledge about environment dynamics and potential safety outcomes. Similarly, while offline RL methods~\cite{levine2020offline, dukkipati2025active} seek to learn performant policies from fixed datasets, they often lack the mechanisms for safety-aware reasoning or formal falsification. To address this gap, we introduce a novel \emph{risk-aware epistemic uncertainty} metric that jointly accounts for insufficient data coverage and for potential safety violations in a structured manner. For example, in the context of Type 1 Diabetes management, where $\pi_{\text{task}}$ recommends insulin doses based on glucose levels and meal intake, standard active learning might prioritize states with high variance in glucose levels (i.e., epistemic uncertainty), while offline RL would focus on improving expected glycemic control (i.e., reward maximization). Our approach instead targets areas where both the epistemic uncertainty and risk of hypoglycemia are elevated—such as post-exercise meals not present in $\mathcal{D}$—allowing targeted exploration or falsification to address critical safety blind spots.

\section{Approach}
This section presents our approach (Fig. \ref{fig:overview}) that combines formal verification with guided simulation-based search in three stages. We (1) abstract the RL policy into a CAPS graph and verify its safety properties using model checking via Storm. If no violations are found, we (2) perform targeted rollouts from high-risk and uncertain states to uncover violations missed by abstraction, and then (3) repair the policy utilizing the detected violating trajectories. Unlike standalone model checking, our method addresses abstraction gaps through targeted exploration; unlike heuristic search methods, it is formally guided by probabilistic risk estimates derived from model checking.

\subsection{Formal Verification via XRL Abstraction}
 Simulation-based falsification methods such as fuzzing and search-based testing~\cite{karakovskiy2012mario, manes2021art} are effective at revealing policy failures. However, they offer no formal guarantees and can be computationally expensive. In contrast, probabilistic model checking (e.g., Storm~\cite{hahn2019storm, wang2020towards}) offers rigorous analysis using logical specifications. However, it faces scalability and fidelity loss resulted from abstraction. Thus, to balance tractability with formal rigor, our framework begins with model checking over an abstracted policy. Common abstraction methods, such as clustering, quotient MDPs~\cite{Biza}, and discretization~\cite{Abstraction}, reduce state complexity but lack interpretability. Instead, we adopt CAPS~\cite{McCalmon}, a global explainable RL method that summarizes the entire policy into an interpretable finite-state graph. This provides a verifier-friendly graph while offering structured explanations for debugging and analysis. CAPS uses CLTree~\cite{cltree} to cluster states from offline trajectories and builds a policy graph $\mathcal{G}_\pi= (V, E)$, where nodes are abstract state clusters and edges are task-policy transitions. Transition probabilities are estimated from $\mathcal{D}$, and nodes are annotated with user-defined boolean predicates.
  
To make the abstraction safety-aware, we augment each node with a learned risk estimate based on the expected cumulative safety cost:
\vspace{-5pt}
\begin{equation}\label{q_risk}
Q_{\text{risk}}(s,a) = \mathbb{E} \left[ \sum_{t=0}^{T} \gamma_{\text{risk}}^t , c(s_t, a_t) \right],
\end{equation}

\noindent where $c(s_t, a_t)$ is the safety cost and $\gamma_{\text{risk}}$ a discount factor. We approximate this using a risk critic $\hat{Q}_{\phi,\text{risk}}$ trained from sampled transitions by minimizing:
\vspace{-10pt}
\begin{multline}
J_{\text{risk}}(s_t, a_t, s_{t+1}; \phi) = \frac{1}{2} \left( \hat{Q}_{\phi,\text{risk}}(s_t, a_t) - \right. \\
\left. \left( c(s_t, a_t) + \gamma_{\text{risk}} \mathbb{E}_{a_{t+1} \sim \pi_{\text{task}}(\cdot|s_{t+1})} 
\left[ \hat{Q}_{\phi,\text{risk}}(s_{t+1}, a_{t+1}) \right] \right) \right)^2
\end{multline}

Each node is labeled safe or unsafe by thresholding $\hat{Q}_{\text{risk}}$. We then use Storm to verify PCTL specifications of the form:

\vspace{-5pt}
\begin{equation}\label{pctl}
\varphi \equiv \mathbb{P}_{\geq p}[\neg \text{unsafe} , \mathcal{U} , \text{goal}],
\end{equation}

\noindent asserting that with probability at least $p$, the goal is reached without visiting unsafe states. If safety specification $\varphi$ is violated, Storm returns a counterexample that identifies a sequence of abstract states leading to any unsafe behavior. 

\begin{definition}[$\varphi$-Counterexample]
Let $\mathcal{G}_\pi = (V, E)$ and $\varphi$ be the specification above. A counterexample trace is a path $\tau = (v_0, \dots, v_T)$ such that: (1) some $v_t \in \text{unsafe}$, and
(2) the probability of satisfying $\varphi$ along $\tau$ is less than $p$.
\end{definition}

\begin{figure}[tb]
    \centering
    \includegraphics[width=0.7\linewidth]{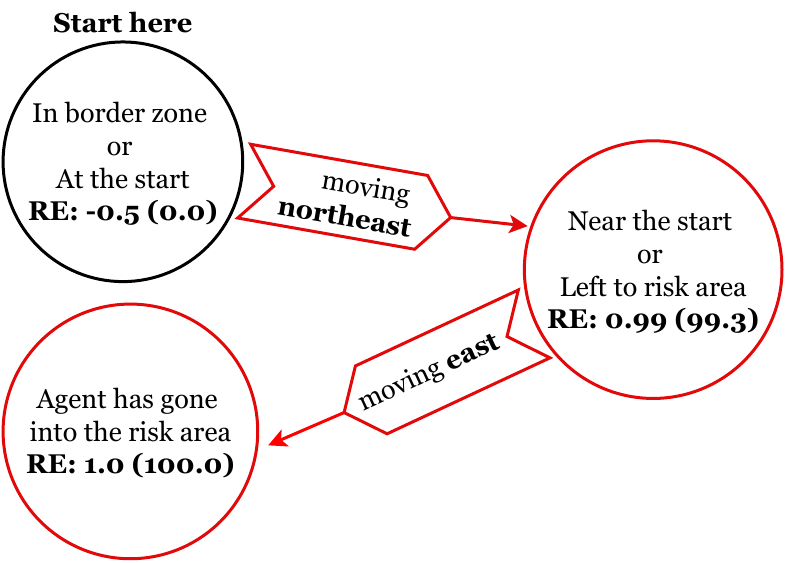}
   \caption{A counterexample detected by Storm then visualized in CAPS in Navigation2 environment highlighted in red. Each node shows a semantic summary, associated risk estimate (RE), and its normalized value in parentheses to aid end-user interpretability.}
    \label{fig:viz1}
    \vspace{15pt}
\end{figure}

\noindent Storm returns such a counterexample trace in $\mathcal{G}_\pi$, annotated with predicates and risk values. This interpretable trace helps developers visualize, diagnose, and repair unsafe behavior (Fig.~\ref{fig:viz1}).

\subsection{Risk- and Uncertainty-Guided Falsification}

When formal verification yields no counterexamples, we cannot conclude with certainty that the policy satisfies the safety property in Eq.~\ref{pctl}. This limitation stems from coverage gaps in the offline dataset, which may fail to include trajectories that expose critical failure modes. To address this, we leverage epistemic uncertainty as a proxy for data coverage: regions of the state space with high uncertainty are assumed to be underexplored and potentially risky. We introduce a falsification strategy guided by this uncertainty, combined with risk estimation, to actively search for safety violations that formal verification may have missed. This procedure is activated only when model checking reports no violation and operates under a constrained rollout budget defined by the number of seed states $N$ and a maximum trajectory depth $L$.

\begin{algorithm}[tb]
\caption{Risk- and Uncertainty-Guided Falsification  \label{alg:byz}}
\textbf{Parameters:} Number of Initial Risky States $N$, Depth Limit $L$, Uncertainty Threshold $\delta$, Environment $e$, Risk Threshold $\alpha$, Policy  $\pi$ and Risk Critic $Q_{\text{risk}}$.
\begin{algorithmic}[1]
\STATE model $\leftarrow$ UncertaintyModel(numModels = 5)
\STATE violatedStates $\leftarrow$ []
\FOR{state $s=1,...,N$ in risky states}
    \STATE currentState $\leftarrow$ s
    \STATE done $\leftarrow$ False
    \STATE violated $\leftarrow$ False
    \WHILE{!done and !violated}
        \STATE $a \leftarrow \pi(\text{currentState})$
        \STATE $\text{uncertainty} \leftarrow \text{model}(a, \text{currentState})$ ~~  (Eq.\ref{eq-uncer})
        \FOR{depth $l=1,...,L$}
            \IF{$\text{uncertainty} \leq \delta$}
             \STATE \text{currentState} $\leftarrow$ mutate(currentState)  ~~  (Eq.\ref{eq-mutate})
                \STATE $a\leftarrow \pi(\text{currentState})$
                \STATE $\text{uncertainty} \leftarrow \text{model}(a, \text{currentState})$
                
            \ELSE
               \STATE break
            \ENDIF
        \ENDFOR
        \STATE risk $\leftarrow$ $Q_{\text{risk}}(\text{currentState},a)$  ~~  (Eq.\ref{q_risk})
        \IF{risk $> \alpha$} 
            \STATE violated = True
            \STATE violatedStates.append(currentState)
        \ENDIF
        \STATE currentState, reward, done $\leftarrow$ e.step(a)
    \ENDWHILE   
\ENDFOR
\end{algorithmic}
\end{algorithm}


\paragraph{Seed Selection and Rollout (Alg.~\ref{alg:byz}, lines 3–4).}
Our falsification procedure begins with the selection of high-risk initial states. Specifically, we rank each candidate state $s \in \mathcal{D}$ using the risk estimate from the model checker and select the top-$N$ states with the highest risk score. For each seed $s_0$, we roll out trajectories using the agent policy $\pi$, evaluating epistemic uncertainty for each state-action pair using an ensemble of $K$ encoders $\mathcal{E}_k$ trained via contrastive learning \cite{dukkipati2025active}. Each encoder computes a joint embedding: 
\[
\mathcal{E}_k(s, a) = \mathcal{E}_k^{\mathbf{s}}(s) + \mathcal{E}_k^{\mathbf{a}}(a),
\]
and uncertainty is quantified as:
\vspace{-5pt}
\begin{equation} \label{eq-uncer}
\text{Uncertainty}(s, a) = \max_{i,j} \| \mathcal{E}_i(s, a) - \mathcal{E}_j(s, a) \|_2^2.
\end{equation} 

\paragraph{Uncertainty Filtering and State Mutation (Alg.~\ref{alg:byz}, lines 8–16).}
If the resulting uncertainty falls below a threshold $\delta$, the state is assumed well-represented in $\mathcal{D}$ and unlikely to reveal novel violations. We then apply a gradient-based mutation to push toward underexplored, risky regions:
\begin{equation}\label{eq-mutate}
s'_t = s_t + \alpha \cdot \nabla_{s_t} \left| Q_{\text{target}}(s_{t+1}) - Q(s_t, a_t) \right|^2,
\end{equation}
where $Q_{\text{target}}(s_{t+1}) = r + \gamma \max_{a'} Q(s_{t+1}, a')$. This mutation process is guided by the observation that high TD-error indicates a significant mismatch between the predicted $Q$ value and the actual return—often signaling that the state is underexplored or it lies in a region where the agent’s learned value function generalizes poorly \cite{fuzz}. Thus, by climbing the TD-error gradient, we generate perturbed states that are likely to be novel or poorly understood, increasing the chance of discovering risky behaviors in regions that has not yet well captured by the offline dataset $\mathcal{D}$.

\paragraph{Violation Detection and Termination (Alg.~\ref{alg:byz}, lines 17–21).}
At each step of the falsification rollout, we evaluate the safety critic $Q_{\text{risk}}$, and if it exceeds a threshold $\alpha$, we flag it as a violation. Otherwise, the rollout proceeds until either a violation is found or the depth limit $L$~ is reached. This falsification method combines risk-aware ranking, epistemic uncertainty estimation, and gradient-guided mutation to efficiently uncover safety violations that model checking may miss—without exceeding interaction constraints.


\subsection{PAC Guarantee for Safety Violation Detection}

We now provide a theoretical guarantee for the completeness of our verification framework. We show that the combination of abstraction-based model checking and risk-uncertainty guided falsification yields Probably Approximately Correct (PAC) safety assurance.

\paragraph{Assumptions}
\begin{enumerate}
    \item \textbf{Abstraction Soundness:} The CAPS graph $\mathcal{G}_\pi$ over-approximates the true reachable space. Any unsafe trajectory $\tau$ in the concrete environment either (i) appears in $\mathcal{G}_\pi$, or (ii) begins in an abstract state flagged as high-risk.

    \item \textbf{Risk Estimator Accuracy:} The learned risk critic $\hat{Q}_{\text{risk}}$ satisfies:
\begin{equation}
\left| \hat{Q}_{\text{risk}}(s,a) - Q_{\text{risk}}(s,a) \right| \leq \epsilon_r \quad \forall (s,a) \in \mathcal{S} \times \mathcal{A},
\end{equation}
where $\epsilon_r > 0$ denotes a uniform upper bound on the estimation error between the predicted and true cumulative safety cost. This ensures that the risk critic provides approximately correct estimates with high probability across all state-action pairs.
    \item \textbf{Uncertainty Estimator Coverage:} The contrastive ensemble assigns high uncertainty to out-of-distribution $(s,a)$ pairs w.r.t $\mathcal{D}$ with failure probability at most $\delta$.
\end{enumerate}

\paragraph{Theorem (PAC Safety Violation Guarantee)}
Let $\mathcal{V}_{\text{found}}$ denote the set of violations detected by Storm and falsification. Under assumptions (1)--(3), the probability of failing to detect any safety violation (if one exists) is at most $\delta$. Formally, the method is $(\epsilon_r, \delta)$-PAC, meaning:
\begin{equation}
\mathbb{P}\left( \exists \, \tau \sim \pi_{\text{task}} : \tau \text{ violates safety and } \tau \notin \mathcal{V}_{\text{found}} \right) \leq \delta.
\end{equation}
\noindent \textit{Proof Sketch}
Let $\tau \in \mathcal{T}{\text{unsafe}}$ be any unsafe trajectory under the task policy $\pi$. We consider two possibilities. First, if $\tau$ exists in the abstracted policy graph $\mathcal{G}\pi$, then, by the soundness of the abstraction (Assumption 1), Storm will detect the violation since $\mathcal{G}\pi$ either contains the unsafe transition explicitly or marks the corresponding abstract state as unsafe. Second, if $\tau$ originates from regions of the environment not represented in $\mathcal{G}\pi$, then by Assumption 3, the ensemble-based uncertainty estimator will assign high epistemic uncertainty to these regions with probability at least $1 - \delta$, prompting their inclusion in the falsification phase. During exploration, if $\tau$ leads to a safety violation, the risk critic $\hat{Q}_{\text{risk}}$, which is $\epsilon_r$-accurate by Assumption 2, will identify at least one risky state-action pair along $\tau$, leading to its detection. Therefore, the probability that an unsafe trajectory $\tau$ both violates the safety specification and remains undetected is at most $\delta$, establishing the PAC guarantee.

\subsection{Safety Shield for Policy Repair}

To mitigate safety violations discovered during formal verification and falsification but without retraining the original task policy like previous methods such as \cite{fuzz,fuzz2}, we introduce a lightweight policy repair mechanism via a modular safety shield. Inspired by shielding techniques in safe RL \cite{alshiekh2018safe,recRL,adv}, our approach \textit{dynamically} switches between the task policy and an auxiliary safe policy based on predicted risk levels. The shield monitors the agent’s actions using the previously trained risk critic $Q_{\mathrm{risk}}(s, a)$. When the estimated risk of an action exceeds a safety threshold $\epsilon$, the shield intervenes and overrides the task action $a_t \sim \pi_{\mathrm{task}}(s_t)$ with an alternative action $a_t^{\mathrm{safe}} \sim \pi_{\mathrm{safety}}(s_t)$ from a dedicated safe policy:
\begin{equation}
    \text{Shield}(s_t, a_t): Q^{\pi_{\mathrm{task}}}_{\mathrm{risk}}(s_t, a_t) > \epsilon.
\end{equation}

\paragraph{Training the Safety Policy.}
To construct $\pi_{\mathrm{safety}}$, we adopt an adversarial regularization approach. We first synthesize an adversarial policy $\pi_{\mathrm{adv}}$ that intentionally seeks high-risk regions by maximizing expected safety costs as in \cite{adv}. The safe policy is then trained to diverge from this adversary, learning to avoid high-risk actions by solving:
\begin{equation}
    \pi_{\mathrm{safe}} = \arg\max_{\pi} \mathrm{KL}(\pi \parallel \pi_{\mathrm{adv}}),
\end{equation}
where $\mathrm{KL}(\cdot \parallel \cdot)$ denotes the Kullback-Leibler divergence. This objective encourages the safe policy to avoid the regions favored by the adversary, thus reducing safety violations.




\paragraph{Visualizing Shield Effects in the Abstraction.}
After applying the safety shield, we regenerate the CAPS graph to reflect policy modifications. Each edge in the updated graph is annotated as either a ``task decision'' or a ``safety decision,'' depending on whether the action was chosen by $\pi_{\mathrm{task}}$ or $\pi_{\mathrm{safety}}$, respectively. The dominant policy at each abstract node $B$ is inferred from the majority of associated concrete actions. This visualization enables RL practitioners to assess where and how the shield intervened, providing actionable insight into the safety-performance trade-off.

\begin{figure}[tb]
    \centering
    \includegraphics[width=0.7\linewidth]{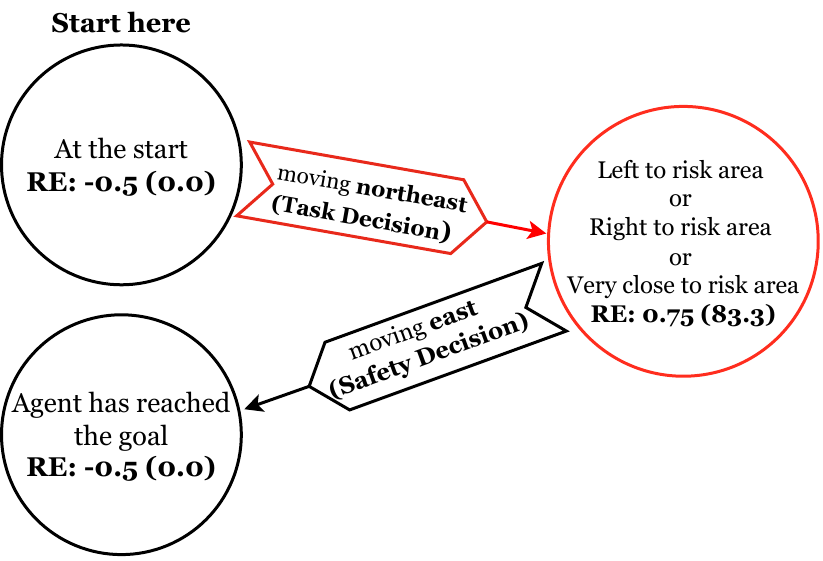}
  \caption{CAPS graph showing shielded policy execution in Navigation2. Edges are labeled by the deciding policy—task or safety—and red highlights high-risk nodes where the shield intervened. Risk estimates (RE) are shown with normalized values in parentheses.}
    \label{fig:shield}
    \vspace{15pt}
\end{figure}

\section{Experiments}
\subsection{Experiment Setup} 

\vspace{3px}
\noindent \textbf{Tasks.} We evaluate our method on three safety-critical RL domains: (1) \textit{Navigation2} and (2) \textit{Maze} from the CMDB Mujoco benchmark~\cite{recRL}, and (3) a simulated \textit{Type 1 Diabetes} insulin dosing task. Navigation2 and Maze feature continuous state-action spaces, where the agent must reach a designated goal while avoiding collisions with obstacles or walls. These tasks are chosen for their geometric complexity and stochastic dynamics, which naturally give rise to diverse modes of failure. The diabetes environment simulates patient glucose dynamics over a 24-hour period with measurements every 10 minutes. The agent selects discrete insulin doses (0 to 10 mU in 0.5 mU increments) based on the patient’s physiological state, including current and past glucose levels, gut absorption, and meal events. The goal is to maintain glucose within the safe range of 65–105 mg/dL.\footnote{We used the simulator from \url{https://github.com/strongio/dosing-rl-gym} to train our agent using the SAC algorithm} This medical task provides a partially observable and high-stakes setting, where safety violations are context-sensitive and may emerge from subtle interactions between physiological variables and agent actions. In all environments, we evaluate well-trained SAC agents~\cite{sac} as the target policies. Falsification experiments are run with a 100-episode rollout budget. Full environment, agent training details, and thresholds are provided in Appendix~A.

\vspace{3px}
\noindent \textbf{Baselines.} We compare our falsification component with two state-of-the-art strategies: (1) \textit{Uncertainty-Based Search}~\cite{dukkipati2025active}, which prioritizes exploration of states underrepresented in the offline dataset using epistemic uncertainty, and (2) \textit{DRLFuzz}~\cite{fuzz}, which expands states based on estimated risk without regard for dataset coverage. These baselines represent two complementary philosophies—uncertainty-driven exploration versus risk-driven targeting—against which we benchmark the combined guidance of our proposed method.

\begin{figure}[tb]
    \centering
    \includegraphics[width=\linewidth]{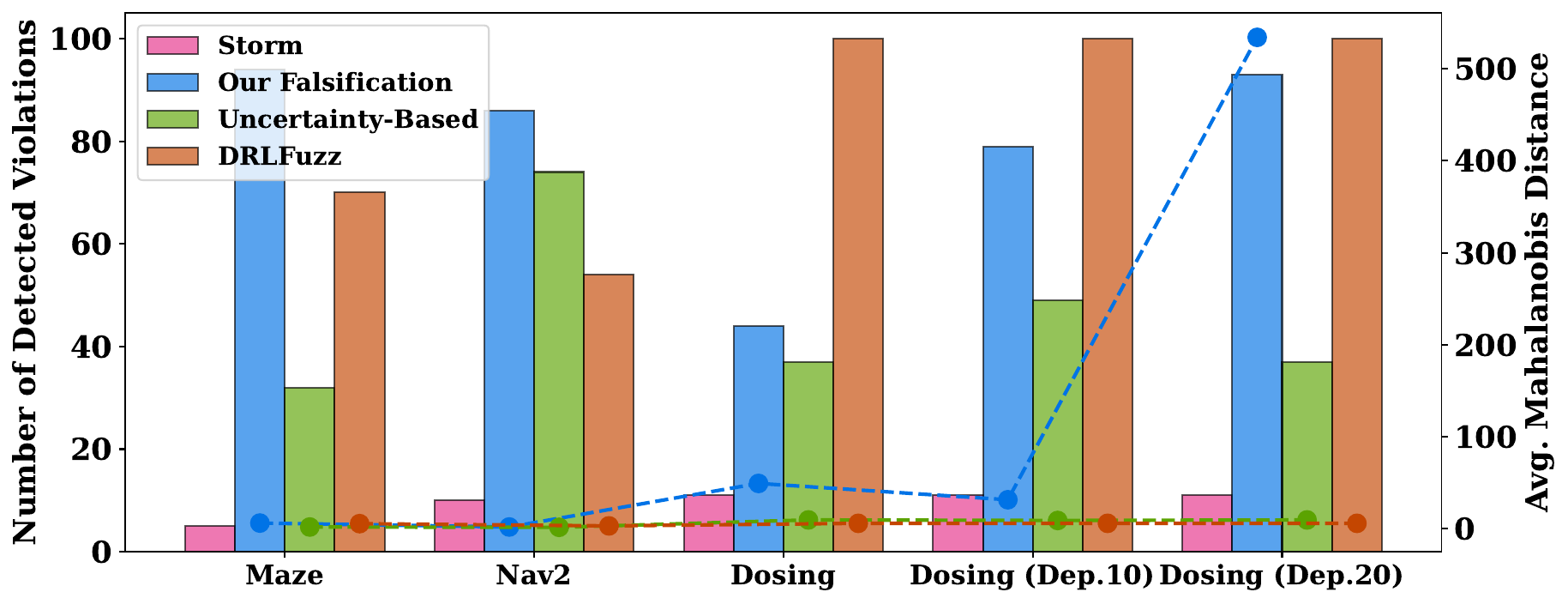}
    \caption{\textbf{RQ1}: Number of detected violations per method across environments, with Mahalanobis distance showing how far the violations are from the dataset. Our framework (Storm + Falsification) consistently finds more violations with higher novelty. }
    \label{fig:RQ1}
    \vspace{20pt}
\end{figure}

\vspace{3px}
\noindent \textbf{Evaluation Metrics and Research Questions.}  
We evaluate our framework along four key axes:  
\begin{itemize}
    \item \textbf{Violation Count (Effectiveness):} The total number of distinct safety violations identified by Storm (model checking) and guided falsification. This measures the overall effectiveness of each method in surfacing unsafe behaviors.
    
    \item \textbf{First Violation Detection (Efficiency):} The number of rollout episodes required to detect the first safety violation during falsification. This reflects how quickly each method can surface a critical failure under a constrained rollout budget.
    
    \item \textbf{Execution Time (Efficiency):} The total wall-clock time required to perform Storm verification and guided falsification, reported separately. This complements episode-based efficiency by capturing the computational cost of each component. All experiments were conducted on a machine with an AMD EPYC Milan CPU (x86\_64 architecture), 8 logical CPUs, 1 physical core per CPU, and 21.6 GB of RAM.
 
    \item \textbf{Violation Diversity:} The mean pairwise distance among latent representations of unsafe trajectories discovered via falsification. Higher values mean greater spatial and behavioral spread in the state-action space, reflecting broader coverage of potential failure modes.
    
    \item \textbf{Violation Novelty:} The average distance between discovered unsafe trajectories and those in the offline dataset, measured in the same latent space and computed via the average Mahalanobis distance. This quantifies the extent to which the detected failures represent novel behaviors not captured during training.
\end{itemize}
For evaluation, we pose several research questions (RQs) below (the tabular results are included in Appendix~B).
\begin{itemize}
    \item \textbf{RQ1:} How effective is the proposed framework in detecting safety violations compared to standalone model checking, our falsification strategy, and baseline methods? And How novel are the violations relative to the trajectories in the offline dataset?
 \item \textbf{RQ2:} How diverse are the violations discovered by our guided falsification strategy? 
    \item \textbf{RQ3:} How efficient is the falsification strategy in terms of both the number of trajectories required to detect the first violation and the overall falsification runtime?
        \item \textbf{RQ4:} What is the impact of using CAPS for abstraction on formal verification performance compared to a clustering-based abstraction using KMeans?
    \item \textbf{RQ5:} How does runtime safety shielding affect the frequency of safety violations under the same verification-falsification framework?
\end{itemize}

\subsection{Results and Discussion}

We apply our falsification strategy even when Storm detects violations to assess overlap and complementarity between verification and search, reporting aggregate results across environments and baselines. As part of an ablation, we vary the mutation depth (state mutation frequency) using values of 1, 10, and 20. Results in Navigation2 and Maze remain consistent, suggesting limited sensitivity to depth due to immediate safety feedback. In contrast, Dosing benefits from deeper mutations, consistent with the domain's delayed physiological responses, where safety violations such as hypoglycemia emerge only after a series of cumulative missteps rather than immediate feedback. We also rerun experiments with a safety-shielded agent to evaluate the effectiveness of runtime policy repair within our framework.

\begin{figure}[tb]
    \centering
    \includegraphics[width=\linewidth]{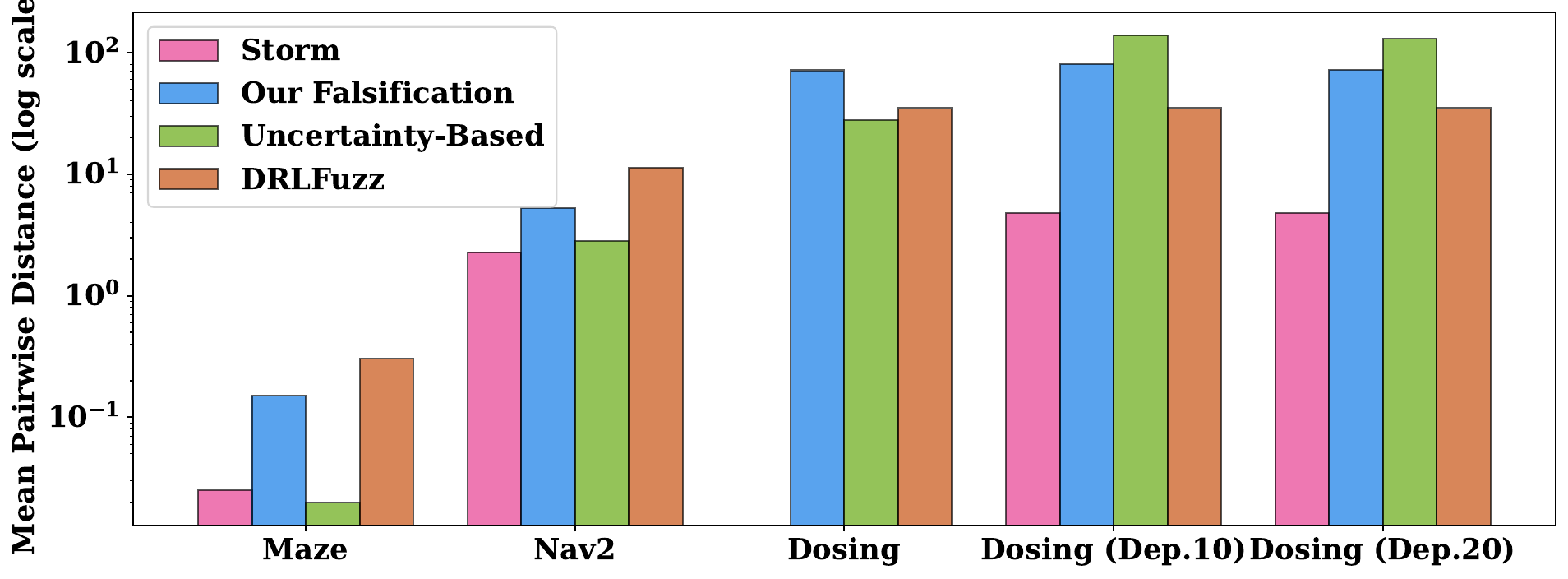}
    \caption{\textbf{RQ2}: Mean pairwise distance of violating trajectories discovered by each falsification method across environments. Higher values indicate more diverse safety violations. Values are plotted on a logarithmic scale to emphasize differences in small-scale environments like Maze and Navigation2.}
    \label{fig:RQ2}
    \vspace{15pt}
\end{figure}

\vspace{5pt}
\noindent \textbf{RQ1 (Fig.~\ref{fig:RQ1}):} This question examines how effective each method is at uncovering safety violations and how different the discovered violations are from those encountered during training. Across all environments, the Storm model checker identified a small number of violations (5–11 trajectories). These are formally verified counterexamples, but their coverage is inherently limited by the fidelity of the abstraction and the coverage of the offline dataset. While our framework terminates after model checking if a violation is found, we executed the falsification stage regardless to assess its added benefit. Our guided falsification approach outperformed both DRLFuzz and uncertainty-based search in Maze and Navigation2, finding 94 and 86 violations, respectively. In the Dosing environment, DRLFuzz achieved the highest violation count (100), though this was accompanied by relatively lower novelty. Specifically, the average Mahalanobis distance of DRLFuzz violations in Dosing (5.8) was far lower than that of our approach, which exceeded 49 in standard depth and reached 534 at depth 20. This distance metric indicates how far the violations are from the distribution of the training data—suggesting that our method detects failures in less explored, higher-risk regions of the state space. In contrast, DRLFuzz, though effective in raw count, tends to revisit more common failure regions. This distinction is especially critical in complex, temporally sensitive domains like insulin dosing, where new failure modes often emerge in rare or sparsely sampled conditions.

\vspace{5pt}
\noindent \textbf{RQ2 (Fig.\ref{fig:RQ2}):}To evaluate the diversity of the detected violations, we measure the mean pairwise distance between the violating trajectories discovered by each method. This metric captures how semantically varied the unsafe behaviors are within each method’s outputs, offering insight into whether the detected violations are close to each other or concentrated in the same regions of the state space. The results in Figure~\ref{fig:RQ2} are plotted on a logarithmic scale to emphasize differences across environments, especially where absolute values are small. In the Dosing environments (standard, depth 10, and depth 20), our guided falsification strategy consistently achieves higher diversity than DRLFuzz and significantly outperforms Uncertainty-Based search in two out of three settings. For example, in the base Dosing task, our method reaches a mean pairwise distance of 71.54, compared to 27.82 for Uncertainty-Based and 34.96 for DRLFuzz. These results demonstrate that our method explores a broader range of unsafe behaviors in complex, real-world environments like medical dosing. However, in simpler domains like Navigation2 and Maze, DRLFuzz occasionally achieves higher diversity than our method. For instance, in Navigation2, DRLFuzz yields a pairwise distance of 11.36 versus 5.23 for our method, and in Maze, its diversity is slightly higher (0.30 vs. 0.15). This suggests that while our risk-guided strategy excels in high-dimensional, structured domains, it may be less effective in simpler environments where some randomness in the search—like that used by DRLFuzz—can help uncover more varied violations. Overall, these results indicate that our approach reliably captures a wide range of unsafe behaviors in complex domains, which is especially important for applications like healthcare. The log-scaled plot highlights that even modest differences in small environments can reflect distinct behavioral patterns, reinforcing the value of evaluating diversity alongside violation count.

\begin{figure}[tb]
    \centering
    \includegraphics[width=\linewidth]{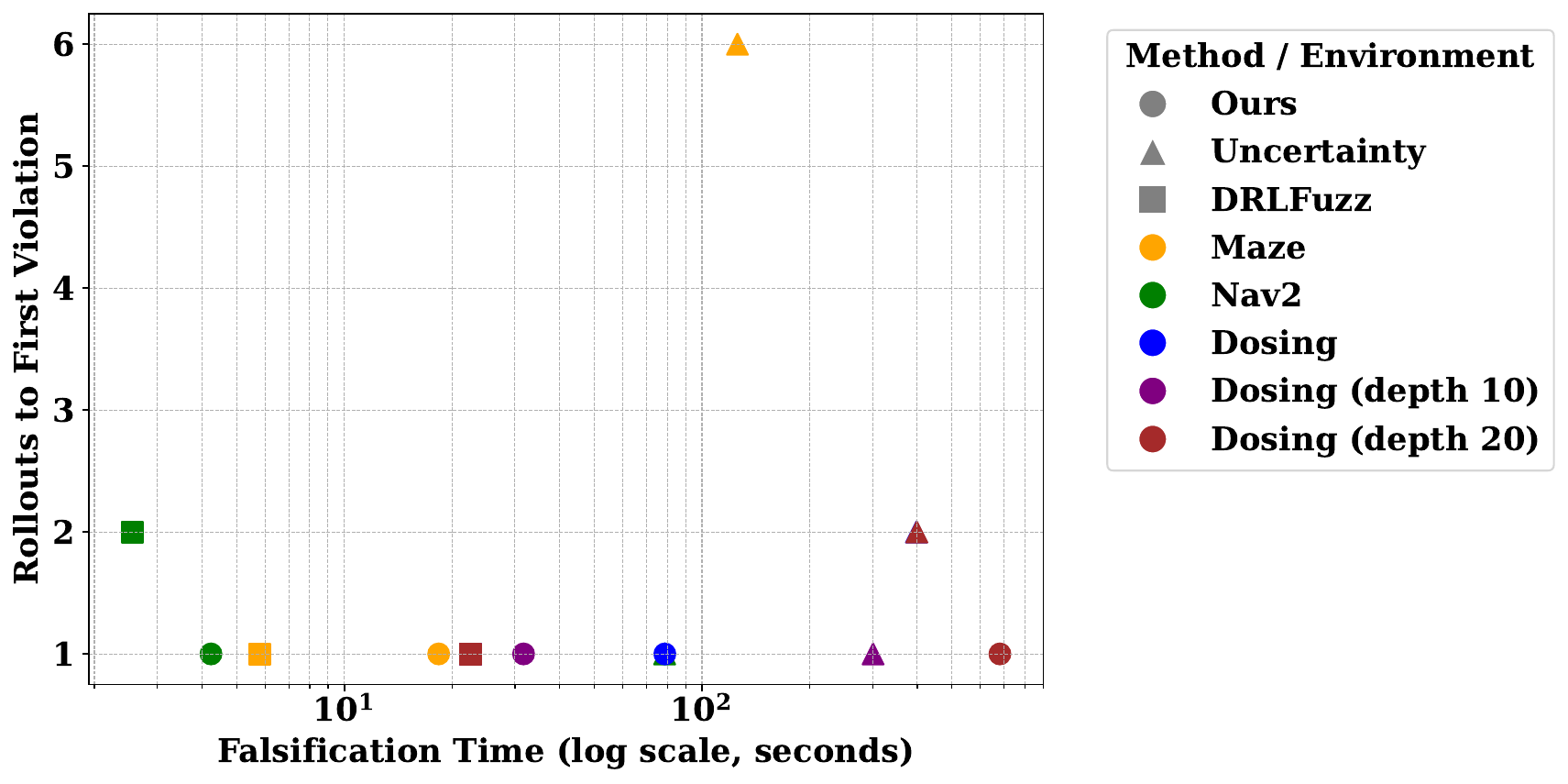}
     \caption{\textbf{RQ3:} Efficiency of falsification methods: rollouts to first safety violation vs. falsification time across environments and methods (log scale on x-axis). Lower values are better on both axes.}
    \label{fig:RQ3}
    \vspace{20pt}
\end{figure}

\vspace{5pt}
\noindent \textbf{RQ3 (Fig.\ref{fig:RQ3}):}
We assess efficiency using two metrics: (1) the number of trajectories required to discover the first safety violation, and (2) the total time taken to perform falsification. As shown in Fig.\ref{fig:RQ3}, our falsification strategy consistently identifies the first violation with minimal rollout cost—requiring only a single trajectory across all environments, including the more complex Dosing tasks. In contrast, the uncertainty-based strategy takes longer in both metrics, especially in Maze and the deeper Dosing scenarios (6 and 2 trajectories respectively), with falsification times reaching up to 398 seconds. DRLFuzz shows strong performance in rollout budget and time, often matching our method in the number of rollouts to the first violation (1 or 2 trajectories) and achieving the lowest execution time overall (e.g., only 2.5s in Navigation2 and ~22s in Dosing). However, as shown in RQ1 and RQ2, this comes at the cost of reduced trajectory diversity (in Dosing) and less effective exploration of underrepresented or high-risk regions. In summary, our falsification approach strikes a favorable balance: it matches DRLFuzz in minimal trajectory requirements, is significantly faster than uncertainty-based methods in all domains except Dosing (depth 20), and—when paired with its stronger violation count and competitive diversity from earlier RQs—demonstrates a strong trade-off between speed, effectiveness, and exploration completeness.

\vspace{5pt}
\noindent \textbf{RQ4:} To assess the impact of abstraction choice on verification performance, we replaced CAPS with a KMeans-based policy abstraction and compared results on violation count and action fidelity across Maze, Navigation2, and Dosing. Although both methods generated comparable numbers of abstract states, CAPS enabled Storm to detect 5–11 violations per environment, while KMeans found only 2 violations in Navigation2 and none in Maze or Dosing. This discrepancy stems from the difference in action fidelity: CAPS preserved the task policy's action in over 83\% of abstract states in Navigation2, while KMeans achieved only 60\%. Similar trends appear in Maze (51\% vs. 36\%) and Dosing (37\% vs. 25\%). Despite the simplicity of KMeans, its poor alignment with the agent’s actual decision-making led to abstractions that do not accurately capture policy behavior. As a result, formal safety analysis using KMeans-based graphs becomes unreliable. These findings underscore that CAPS not only enhances interpretability but also maintains fidelity crucial for trustworthy verification.

\begin{figure}[tb]
    \centering
    \includegraphics[width=\linewidth]{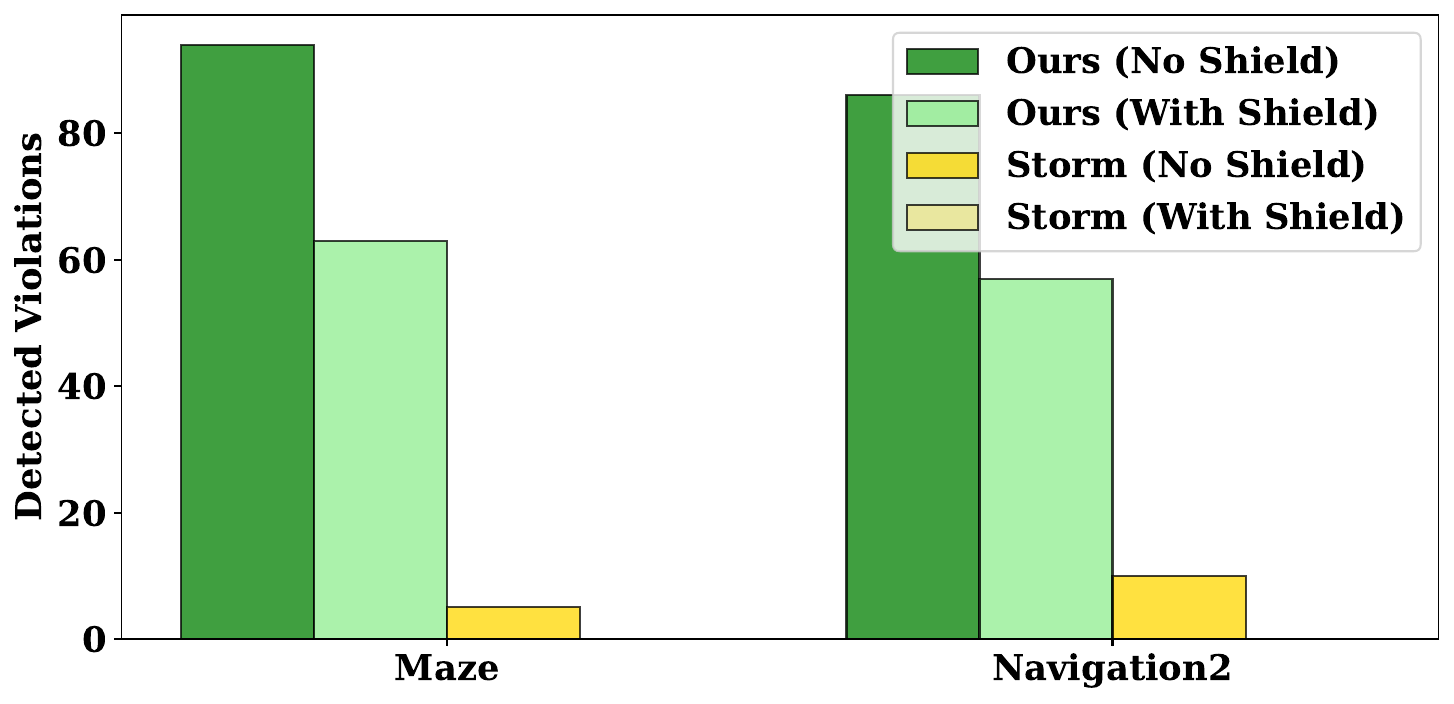}
    \caption{\textbf{RQ5:} Impact of safety shielding on detected violations in Maze and Navigation2. After shielding, Storm detects no violations, while falsification finds fewer failures—highlighting both the shield’s effectiveness and the continued need for falsification to ensure safety.}
    \label{fig:RQ5}
    \vspace{15pt}
\end{figure}

\vspace{5pt}
\noindent \textbf{RQ5 (Fig.\ref{fig:RQ5}):} We report the number of violations detected by Storm and by our falsification strategy with and without shielding in the Maze and Navigation2 environments. In both environments, Storm fails to detect any violations once the shield is active, despite the presence of remaining violations found via falsification. This shows that shielding may help the policy mistakenly pass formal safety specifications, which supports our design of falsification as a necessary second line of verification when the model checker fails to flag unsafe behavior. Our falsification reveals that shielding reduces the number of detected violations by 33\% in Maze (from 94 to 63) and 34\% in Navigation2 (from 86 to 57). These results show that the shield effectively blocks risky policy decisions identified at runtime by the risk critic. We omit shielding results for the Dosing environment as its safety constraints are not directly encoded in the environment’s dynamics or terminal conditions as in CMDP benchmarks of Maze and Navigation2. Violations in this domain (e.g., hypoglycemia) are temporally extended and depend on delayed physiological effects, which makes training an effective runtime shield non-trivial. This limitation arises from the domain design—not from our framework—and motivates future work on designing temporally-aware shields for safety-critical medical applications.

\section{Conclusion}  
We addressed the challenge of verifying and falsifying safety properties in reinforcement learning policies under limited data coverage by designing a hybrid framework that combines explainable abstraction, probabilistic model checking, and risk-aware falsification. While effective in diverse safety-critical tasks, our approach has two main limitations. First, CAPS does not currently support image-based observation spaces; future work will explore extending CAPS or developing new abstractions for high-dimensional inputs. Second, we omit shielding results for the Dosing task, as its temporally extended safety violations and partially observable dynamics make runtime repair nontrivial. This motivates future research on designing temporally-aware shields tailored to medical domains with delayed effects and implicit constraints.

\bibliography{ECAI_main}

\end{document}